\documentclass[review]{elsarticle}

\usepackage{lineno,hyperref}
\usepackage{pdfpages}
\usepackage{amsfonts}
\usepackage{amsmath}
\usepackage{threeparttable}
\usepackage{booktabs}
\usepackage{multirow}
\usepackage{array}
\usepackage{caption}
\usepackage{float}
\usepackage{subfigure}
\usepackage{amssymb}
\usepackage{graphicx}
\usepackage{url}
\usepackage{makeidx}

\journal{arXiv.org}

\begin{document}

\begin{frontmatter}

\title{GPU Accelerated Cascade Hashing Image Matching for Large Scale 3D Reconstruction}

\author{Tao Xu}
\author{Kun Sun}
\author{Wenbing Tao\corref{cor1}}
\ead{wenbingtao@hust.edu.cn}
\address{School of Automation, Huazhong University of Science and Technology, Wuhan 430074, China}
\cortext[cor1]{Corresponding author.}





\begin{abstract}
Image feature point matching is a key step in Structure from Motion(SFM). However, it is becoming more and more time consuming because the number of images is getting larger and larger. In this paper, we proposed a GPU accelerated image matching method with improved Cascade Hashing. Firstly, we propose a Disk-Memory-GPU data exchange strategy and optimize the load order of data, so that the proposed method can deal with big data. Next, we parallelize the Cascade Hashing method on GPU. An improved parallel reduction and an improved parallel hashing ranking are proposed to fulfill this task. Finally, extensive experiments show that our image matching is about 20 times faster than SiftGPU on the same graphics card, nearly 100 times faster than the CPU CasHash method and hundreds of times faster than the CPU Kd-Tree based matching method. Further more, we introduce the epipolar constraint to the proposed method, and use the epipolar geometry to guide the feature matching procedure, which further reduces the matching cost.
\end{abstract}


\end{frontmatter}

\section{Introduction}
Image feature matching is to find the correspondence between images, which is widely used in image mosaic, object detection and 3D reconstruction. SIFT\cite{Lowe2004} is the most often used image feature since it has strong anti-interference ability and high matching accuracy. However, with the development of the Internet and information technology, the amount of image data is getting larger and larger in many computer vision applications. Especially in large scale Structure from Motion(SFM)\cite{Agarwal2009,Crandall2011}, image matching occupies the major computational cost because tens of thousands of images need to be handled. Therefore, fast image matching method is extremely important for accelerating the entire process of 3D reconstruction.

In 2014, Cheng \emph{et al.} proposed a fast image feature matching method based on Cascade Hashing in 3D reconstruction \cite{Cheng2014}. The Cascade Hashing (CasHash) uses the Locality Sensitive Hashing \cite{Charikar2002} (LSH) to quickly determine and reduce a candidate point set, so CasHash avoids fully calculating Euclidean distance between all possible points. Compared with many other algorithms, such as Kd-Tree and LDAHash, CasHash has a quite low time complexity. As reported, on the same CPU computing platform, CasHash is able to achieve slightly better performance against Kd-tree and LDAHash in most cases.

Although CasHash is very fast for image matching, it is still time-consuming when the number of image pairs to be matched is extremely large. For example, for exhaustive matching of Dataset Rome16K \cite{Cornell}, there are 115 million image pairs need to be matched and it will spend half a year by CasHash matching algorithm. Although Bag of Feature \cite{Lazebnik2006} and Vocabulary Tree \cite{Nister2006Robust} indexing are used to reduce the matching list to 30 neighbors, it still costs about 9 hours. Therefore it is necessary to further improve the matching efficiency. With the rapid development of hardware, GPU card is becoming more and more popular. It has been extensively used for high-performance and massively parallel computation. Considering the independence of the matching between different images, which is very suitable for parallelization, we intend to implement GPU-based CasHash to accelerate the matching procedure in large scale SFM 3D reconstruction in the work. However, there are some difficulties that need to be overcome when parallelizing CasHash:

(1) For large scale 3D reconstruction, massive image pairs need to be processed. Limited by memory and GPU memory, the whole data can't be loaded at once. So we need to design an efficient partially data loading and exchanging strategy, which also minimizes the data exchange cost at the same time. (2) In CasHash, there are many reduction (such as calculation of hashing code) and sorting (such as hashing ranking) operations which are the most time consuming parts of Cashash. In parallel computing, reduction and sorting are inefficient operations but can be specialized rewritten for CasHash. 

In order to solve these problems, we propose a GPU accelerated Cascade Hashing Image Matching algorithm for large scale 3D reconstruction. A preliminary version of this work appeared in \cite{Tao2017}. Firstly, a Disk-Memory-GPU data exchange strategy is proposed to optimize the load order for massive images. Secondly, an improved parallel reduction method is proposed to calculate hashing code which is used to determine and reduce a candidate point set. Then we use parallel hashing ranking to select a few nearest neighbors. Finally, the improved parallel reduction method is also used to calculate Euclidean distance between the nearest neighbors and copy output asynchronously.
\section{Related Work}
The development of feature descriptors has made a great contribution to the performance of image matching. In 1999, David Lowe proposed SIFT(Scale-invariant feature transform) feature \cite{Lowe1999Object} and concluded in 2004 \cite{Lowe2004}. SIFT feature is represented by a 128 dimensional vector and uses the Euclidean distances for similarity measurement. In 2004, Yan Ke \emph{et al.}\cite{Ke2004PCA} proposed PCA-SIFT to reduce the complexity of matching. PCA-SIFT use PCA(Principal Component Analysis) to reduce the SIFT feature descriptor from 128 dimension to 20 dimension, which reduces the memory occupied and improves the efficiency of the matching. In 2004, Mikolajczyk \emph{et al.}\cite{mikolajczyk2005a} proposed a variant of SIFT descriptor called GLOH(Gradient Location-Orientation Histogram) that uses log-polar bins instead of square bins to compute orientation histogram. GLOH improves the matching accuracy greatly, but it takes more time to extract features from images. In 2006, Herbert Bay \emph{et al.}\cite{bay2006surf} proposed SURF(Speeded Up Robust Features) descriptor that can improve the robust and speed of feature extract, but SURF is affine invariant under low angle. Based on SURF, Yanwei Pang \emph{et al.}\cite{Pang2012} proposed a fully affine invariant SURF algorithm with high computation efficient at the same time. In recent years, there are also some binary feature descriptors adapted to embedded systems and mobile platforms because of low memory usage and small computational complexity. In 2010, Calonder \emph{et al.} \cite{calonder2010brief} introduced the BRIEF(Binary Robust Independent Elementary Features) descriptor that relies on a relatively small number of intensity difference tests to represent an image patch as a binary string. There are also other binary features such as BRISK\cite{leutenegger2011brisk}, FREAK\cite{alahi2012freak} and etc.

Despite the improvement of descriptors, a series of SIFT feature matching methods are proposed in order to accelerate the speed of image matching. Based on data structure of the binary tree, Kd-Tree \cite{Muja2009} is one of the ANN methods. The time complexity of Kd-Tree is $O\left( N\log N\right)$ \cite{Cormen2009}, which is much faster than brute search. To solve the problem of high-dimensional SIFT feature points, LDAHash \cite{Strecha2012LDAHash} was proposed based on the data structure of hashing. The high 128 dimensional vectors can be reduced into binary hashing codes. And then LDAHash uses the hashing codes as the key to find the nearest neighbors. CasHash is inspired by LDAHash and has more advantages, such as coarse-to-fine search strategy and etc. It makes CasHash faster and resistent to noise point pairs.  Alhwarin \emph{et al.} \cite{Alhwarin2010} proposed VF-SIFT that extend SIFT feature by a few pairwise independent angles. Then this method classifies SIFT features based on their introduced angles into different clusters. Only SIFT features that belong to same cluster are compared. 

On the other hand, in Structure from Motion, there are some other methods to reduce the matching time by changing exhaustive matching into guided matching. Vocabulary Tree Clustering \cite{Nister2006Robust} uses a hierarchically quantized cluster tree to enable scalable clustering and recognition of millions of images. Recognition or classification is performed by running down the tree with each image feature and scoring based on branch traversal. Instead of exhaustive matching, guided matching can be carried out between the images with high similarity score. However, Vocabulary Tree Clustering may miss a small part of correct matches. Li \emph{et al.}\cite{Li2008Modeling} presents an approach that clusters images based on low-dimensional global appearance descriptors, and the clusters are refined using 3D geometric constraints. Thus this method reduces the matching pairs. Each valid cluster is represented by a single iconic view, and the geometric relationships between iconic views are captured by an iconic scene graph. Wu \emph{et al.}\cite{Wu2013} introduce a preemptive feature matching strategy that can reduce the matching pairs by up to $95\%$ while still recovering sufficient good matches for reconstruction. For each image pair (parallelly), it do the following:(a) Match the top-scale features of the two images.(b) If the number of matches from the subset is small, return and skip the next step.(c) Do regular matching and geometry estimation. In this paper, we use GPU accelerated geometry estimation method to calculate fundamental matrix in parallel instead of using the classical RANSAC algorithm. Michal Havlena \emph{et al.}\cite{Havlena2014} proposed VocMatch to establish point correspondences between all pairs of images by a sufficiently large visual vocabulary. Instead of matching each individual image pair, VocMatch directly gets the reconstruction trajectory.

In 3D reconstruction of large scale urban scenes, millions of images need to be processed. Even with the state-of-the-art algorithms, it takes large time to complete 3D reconstruction. As in the paper ''Building Rome in a Day''\cite{Agarwal2009}, Sameer Agarwal \emph{et al.} carried out the experiment that reconstructing cities consisting of 150K images spends less than a day on a cluster with 500 compute cores. With the development of computing platforms and devices, especially CUDA and GPU, a single super PC with GPU cards can achieve similar or better performance than a cluster of CPUs, while the cost is much smaller \cite{Lee2010}. Jan-Michael Frahm \emph{et al.} parallelize the processing of 3D reconstruction on modern multi-core CPUs and GPU cards. This paper processed 3 million images to reconstruction within the span of a day on a single PC.

There are also some parallelizing methods further reducing the matching time. Wu \emph{et al.}\cite{Wu2013} also parallelize the brute force matching on GPU called SiftGPU matching. Shah \emph{et al.} \cite{Shah2015} proposed a parallelizing matching method mainly based on parallelizing Kd-tree. So combine GPU with low time complexity algorithm, the matching time will be reduced greatly.
\section{Overview}
\subsection{Review of Cascade Hashing}
CasHash \cite{Cheng2014} uses two layers of LSH for efficient nearest neighbor search in a coarse-to-fine manner. LSH \cite{Charikar2002} is a kind of ANN method, whose average time complexity is $O\left( 1\right)$. Through hashing mapping, true neighbors are more likely to be assigned to the same bucket, that is, the same hashing code. It consists of the following steps.

\textbf{Step 1. Hashing lookup with multiple tables}\quad All image feature points are mapped into $m$-bit short hashing codes by $m$ different LSH hashing functions. However, using only one hashing table will lead to comparatively large possibility that the real matching pair falls into different buckets. So the above hashing is carried out multiple times, returning multiple lookup tables. Each query point in image $I$ will treat all the points with the same hashing code in image $J$ as its matching candidate.

\textbf{Step 2. Hashing remapping}\quad The query and all the candidate points are mapped into $n$-bit$\left(n>m\right)$ hashing codes with LSH. Since these hash codes are longer than previous ones, the number of points falling into the same bucket with the query is quite small. CasHash computes the Hamming distances between the query point and its candidates.

\textbf{Step 3. Hashing ranking}\quad The Hamming distance between the query and the candidates after hashing remapping ranges from 0 to $n$. CasHash then collects the top $k$ points with the smallest Hamming distance. These points are treated as the final matching candidates of the query point.

\textbf{Step 4. Euclidean distance calculation}\quad Finally, the Euclidean distances of the SIFT feature descriptors between the query point in image $I$ and its $k$ nearest neighbors on image $J$ are computed. Then the point with the smallest distance which also passes Lowe's ratio test \cite{Lowe2004} is selected as the matching point.
\subsection{GPU accelerated Cascade Hashing Image Matching}
\begin{figure}
	\centering
	\includegraphics[width=120mm]{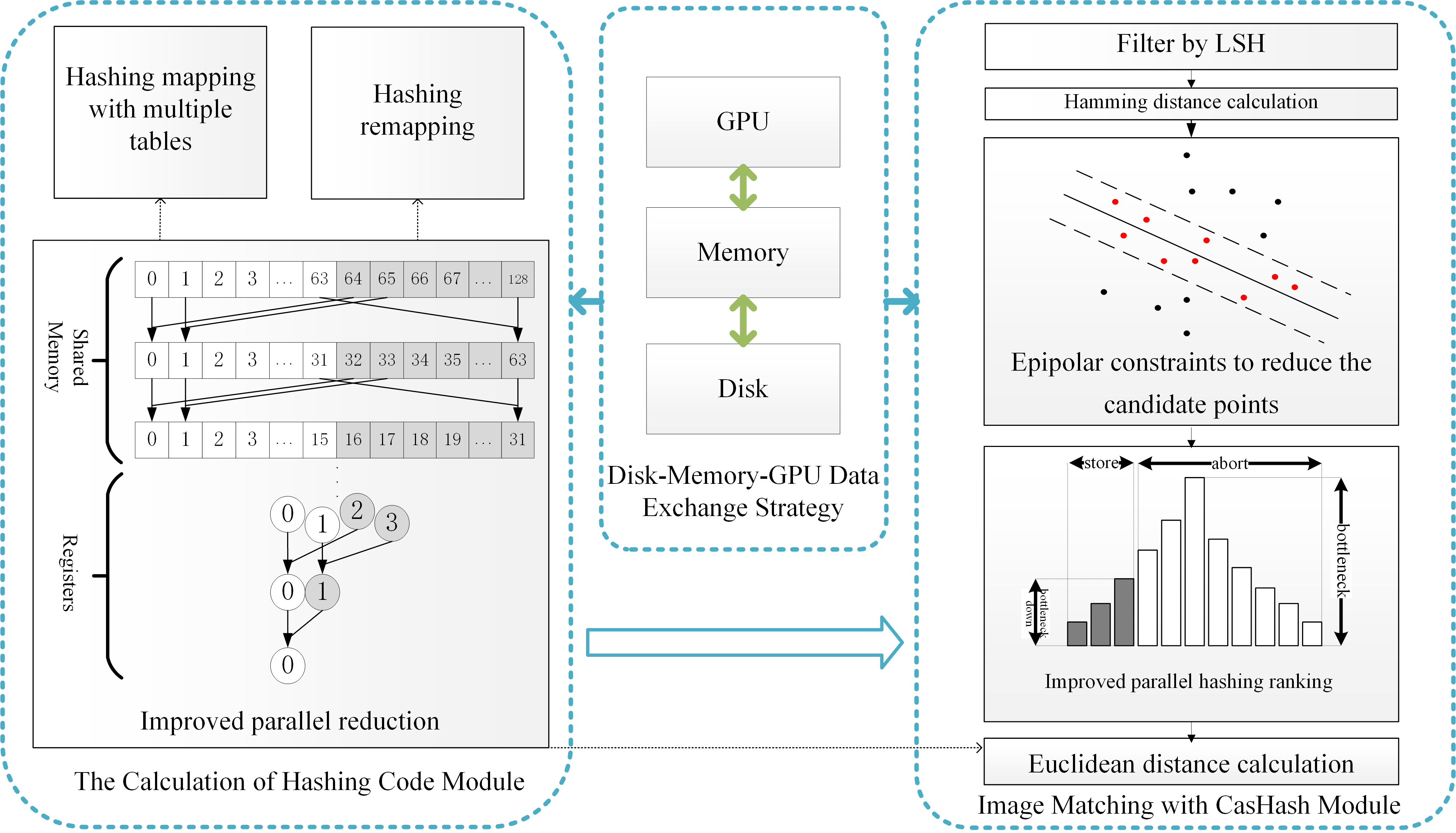}
	\caption{The flow chart of our proposed approach. It contains two modules: hash code calculating module and image matching module. An efficient data scheduling strategy between the disk, memory and graphics card is applied to adapt to large scale data.}
	\label{fig:flow}
\end{figure}

In this paper, we propose a GPU accelerated Image Matching with Cascade Hashing. The flow chart of our method is shown in Fig.\ref{fig:flow}. Firstly, we propose an improved parallel reduction on GPU. By making full use of shared memory and registers on GPU, we use registers to do the last $N_{r}$ rounds of the reduction instead of using shared memory in every round. In step 1, 2 and 4, the calculation of hashing code and Euclidean distance can make use of the improved parallel reduction for acceleration.

In 3D reconstruction of large scale scenes, there are tens of thousands of images to be processed. In this case, the memory usage is easy to exceed the capacities of the main memory and GPU memory. We propose a Disk-Memory-GPU data exchange strategy. A certain number of images are bundled into data blocks, and a certain number of blocks are bundled into data groups. And then we optimize the load order to avoid redundant the data exchange.

In step 3 of CasHash, in order to reduce the computational bottleneck in hashing ranking, we propose a improved parallel hashing ranking which use the threshold $\tau$ to filter the points in the middle of a few buckets. The computational bottleneck can be reduced because the points concentrating in the middle of a few buckets are filtered.

Further more, inspired by the preemptive feature matching strategy \cite{Wu2013}, we use top-scale SIFT features to do exhaustive image matching and compute the epipolar geometry for image pairs. Then the geometric information is used to guide the remaining matching.
\section{Three-Level Data Scheduling Strategy}

Most computational tasks of the proposed method are performed on the graphics card. However, when the number of images scales up, it is impossible to load all the data from the disk to memory and graphics card. Thereby in this section we propose a three-level data scheduling strategy, which satisfies the following two constraints: 1) loading data within the capacity of each device and 2) making full use of the computational units.

Generally speaking, the capacities of disk, memory and graphics card decrease in order. So we organize the whole data in three different granularities accordingly: the whole set, groups and blocks. As is shown in the last row of Fig. \ref{fig:three1} and Fig. \ref{fig:three}, a block contains data from $N_{p}$ images and such $M$ blocks constitute a group. The whole data set consists of $N$ groups and is stored on the disk. For clarity, we will denote $G_i$ as the $i-th$($i\in [1,N]$) and $B_i^j$ as the $j-th$($j\in[1,M]$) data block in $G_i$. In the following part we will introduce two slightly different data scheduling strategies for hash code calculation and image matching because the former operates on a single image while the latter operates between two images.
\subsection{Data Scheduling Strategy for Hashing Code Calculation}

\begin{figure}
	\centering
	\includegraphics[width=0.9\linewidth]{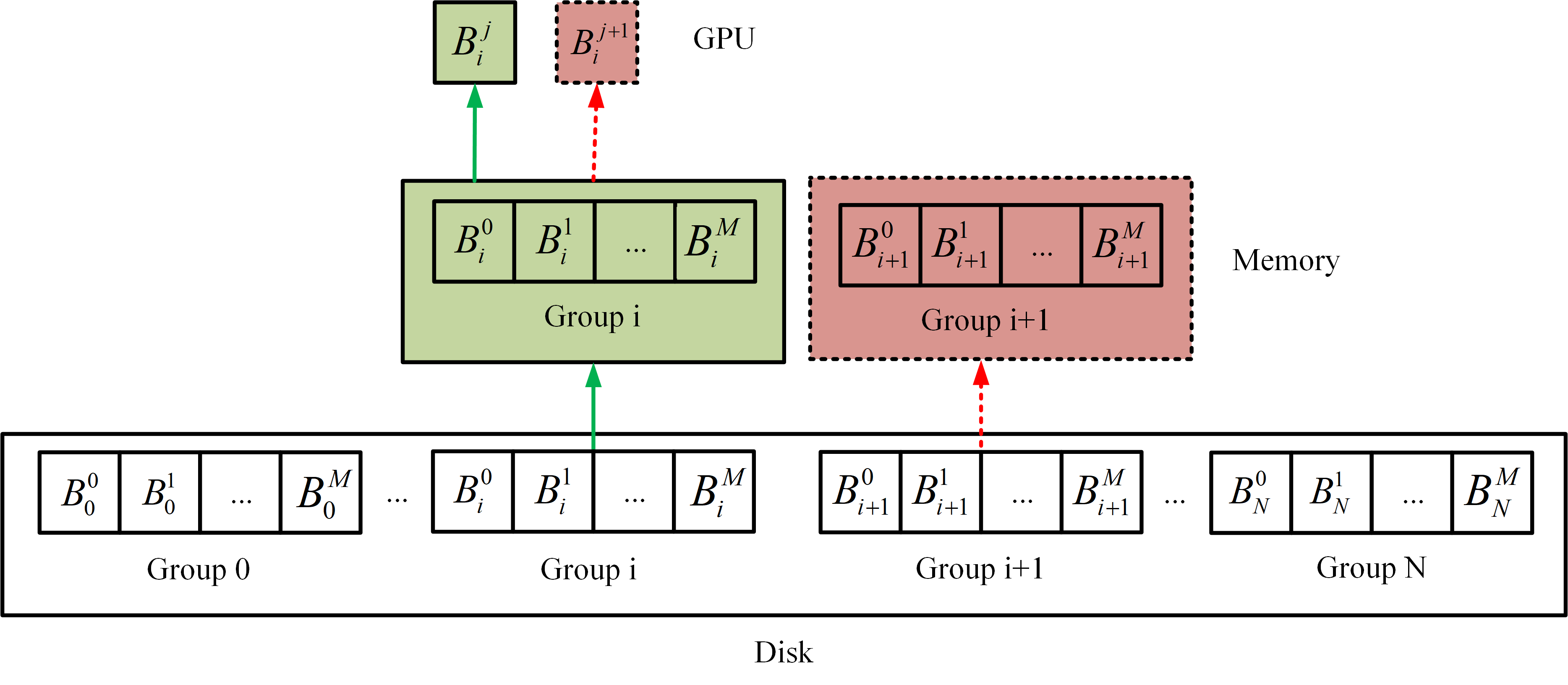}
	\caption{The diagram of the data scheduling strategy when calculating hash code. The solid green rectangle indicates the group and block in progress, while the dashed red rectangle indicates the group and block loaded in advance.}
	\label{fig:three1}
\end{figure}

When calculating hash code we just need to progressively load data to the graphics card in two lines. The first line loads group $G_i$ from the disk to the memory, and then loads one of its un-processed block $B_i^j$ from memory to the graphics card. It is shown as the green arrows in Fig. \ref{fig:three1}. The block $B_i^j$ will be replaced with the next un-processed block $B_i^{j+1}$  if all the data in it has been processed. Similarly, the group $G_i$ will be replaced with another group $G_{i+1}$ if all its blocks have been processed. The second line runs in parallel with the first line on another thread, which is shown as the red arrows in Fig. \ref{fig:three1}. It loads $G_{i+1}$ from the disk when $G_i$ is still in memory, and loads $B_i^{j+1}$ from memory when $B_i^j$ is in the graphics card.

Note that there are two blocks/groups in the graphics card/memory at the same time, but only one of them is in progress. Although $G_{i+1}$ and $B_{i+1}^1$ are coexistent with $G_i$ and $B_i^1$, they will not be processed until $G_i$ and $B_i^1$ are finished and released. Keeping the next group or block loaded in advance ensures that there is no need to wait for new data after the current data is finished, which is more efficient.

\subsection{Data Scheduling Strategy for Image Matching}
\begin{figure}
	\centering
	\includegraphics[width=100mm]{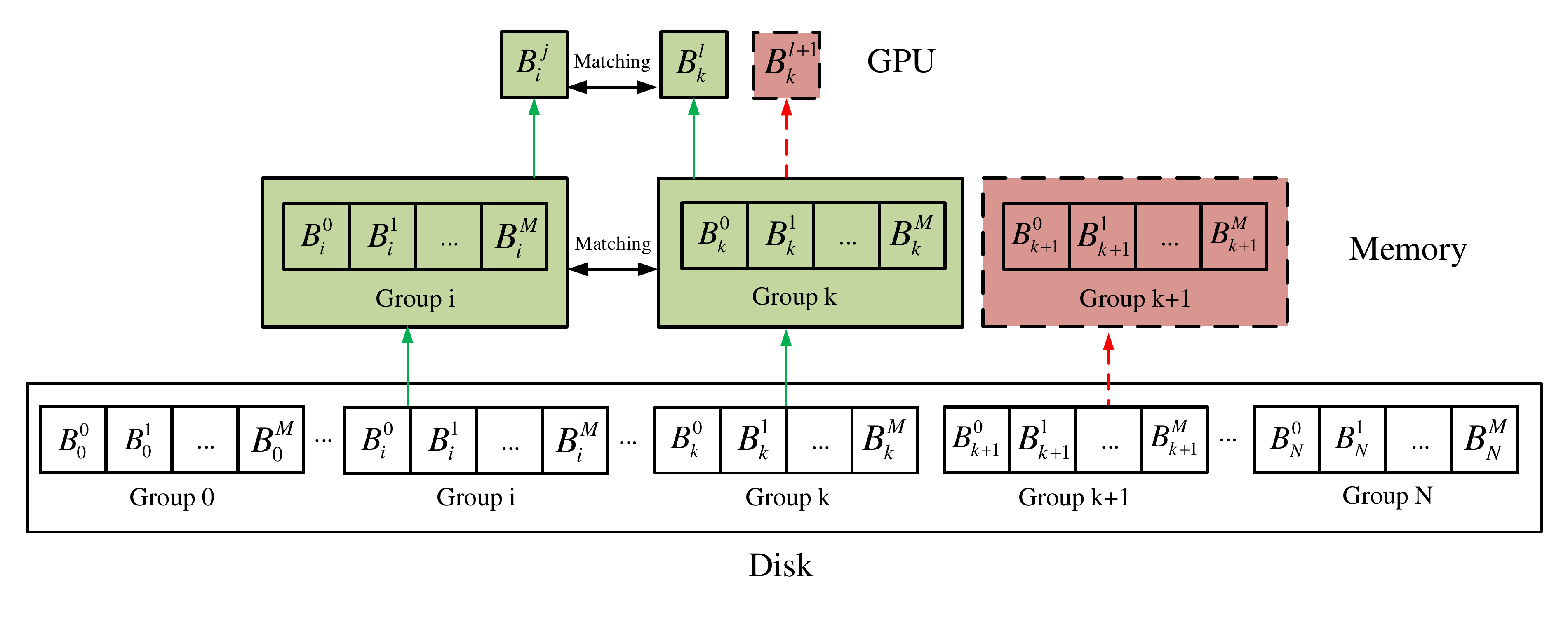}
	\caption{The diagram of the data scheduling strategy when matching images. The solid green rectangles indicate the group and block pairs in progress, while the dashed red rectangle indicates the group and block loaded in advance.}
	\label{fig:three}
\end{figure}

Since image matching operates on a pair of images, the data scheduling strategy becomes more complex. In order to cover all possible image pairs, we need to match all possible group pairs and all possible block pairs. As is shown in Fig.\ref{fig:three}, the strategy here has two lines as well. The first line loads a pair of groups $G_i$ and $G_k$ from the disk to the memory, and then loads one block from each group $B_i^j$ and $B_k^l$ to the graphics card. Data exchanges more frequently between the graphics card and memory than between the disk and memory. After images in $B_k^l$ have been fully matched to images in $B_i^j$, it will be replaced with another new block $B_k^{l+1}$. If $B_i^j$ has been matched to all the blocks in $G_k$, we match images within $B_i^j$ itself and then replace $B_i^j$ with the next block $B_i^{j+1}$. The same rule applies to groups, too. After all the blocks in group $G_k$ have been matched to all the blocks in $G_i$, it will be replaced with another new group $G_{k+1}$. If $G_i$ has been matched to all the other groups, we match blocks within $G_i$ and then replace $G_i$ with the next group $G_{i+1}$. For the same reason described in the previous subsection, the second line which runs on another thread loads $B_k^{l+1}$ and $G_{k+1}$ in advance while existing data is in progress.

The difference with previous subsection is that we focus on pairs rather than individuals. There are three blocks/groups maintained in the graphics card/memory at the same time, and two out of three are in progress. Matching is carried out not only between different blocks/groups, but also within the same block/group.

\section{Fast Computation of Reduction on GPU}
\subsection{The Improved Parallel Calculation of Hashing Code}
In \cite{Charikar2002}, Charikar defined the hashing function by using inner product similarity. In step 1 and step 2 of CasHash, when calculating the hashing code of every SIFT point, we need to compute the inner product of the high dimensional vector. Let $d$ be the dimension of the query vector, so the CUDA kernel function of the hashing map is $Hashing\_kernel<<<N*n,d>>>(PR, PQ)$, where $PR$ and $PQ$ are the pointers of random vectors $r_{k}$ and SIFT feature descriptors in the GPU global memory, respectively. Since the SIFT features point is 128 dimensional, i.e.$d=128$. $N$ is the number of points, and $n$ is the length of hashing code. The kernel function totally calls $N*n$ blocks and each block calls $d$ threads. The inner product vector needs to do $d$ accumulation. Since the GPU global memory access is relatively slow and GPU needs to access GPU global memory repeatedly in the accumulation step, we use the GPU shared memory to reduce access delay \cite{Hong2009}. The shared memory is a kind of GPU cache. The products are stored in the shared memory temporarily when doing accumulation, thus the arithmetic speed is improved.

There are 128 threads participating in operation at the same time in each block. Each thread reads a component of $PR$ and a component of $PQ$, and then multiplies them on shared memory. Data access in shared memory is very fast, but using serial accumulation in each block will activate only one thread, which will reduce arithmetic efficiency. The way to deal with vector accumulation in general parallel computing is parallel reduction \cite{Nickolls2008}, which is also calculated in the shared memory.
\begin{figure}
	\centering
	\includegraphics[width=90mm]{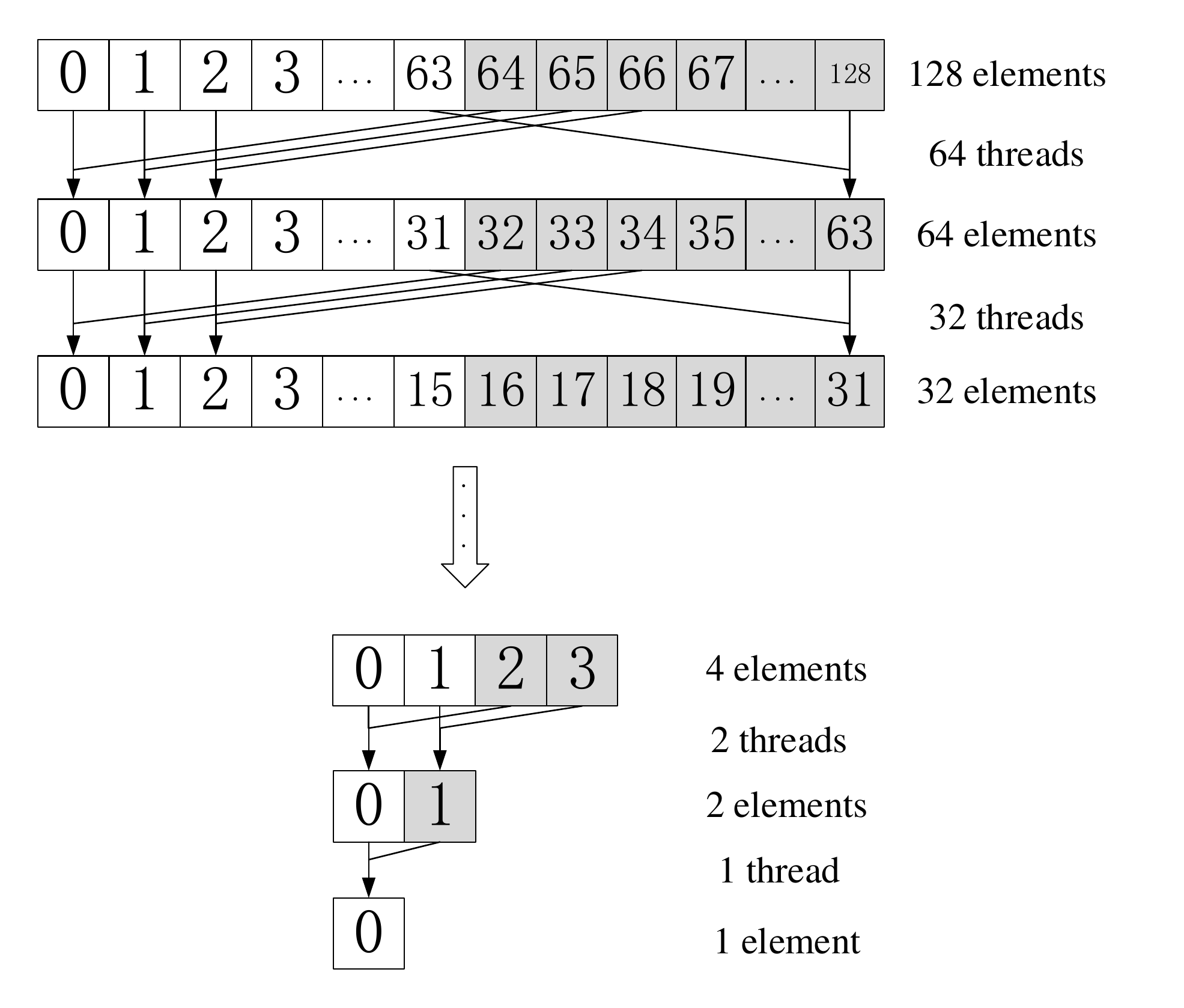}
	\caption{The diagram of general parallel reduction on GPU}
	\label{fig:classic}
\end{figure}

The classic GPU parallel reduction is shown in Fig.\ref{fig:classic}. The accumulation of each block added in the shared memory makes full use of threads. For example, in the first round, the accumulation of 128 elements calls 64 threads, and then obtains 64 results. Consequently, the numbers of elements and threads are reduced by half after every round of operation until the last round, which uses one thread to add two elements. Accumulation is a process of reducing the number of threads used.

The access time for GPU global memory, GPU shared memory and GPU registers are hundreds of operational cycles, 10 operational cycles and a single operation cycle \cite{Cook2012}, respectively. So if registers instead of shared memory is used in the parallel reduction, it can further reduce the delay. However, all the registers could only be accessed by a single thread. If we use the registers in the early stage of parallel reduction, most threads will be idle. In this paper, we improve the parallel reduction by using both the shared memory in the early stage and the registers in the final stage to achieve faster speed.
\begin{figure}
	\centering
	\includegraphics[width=100mm]{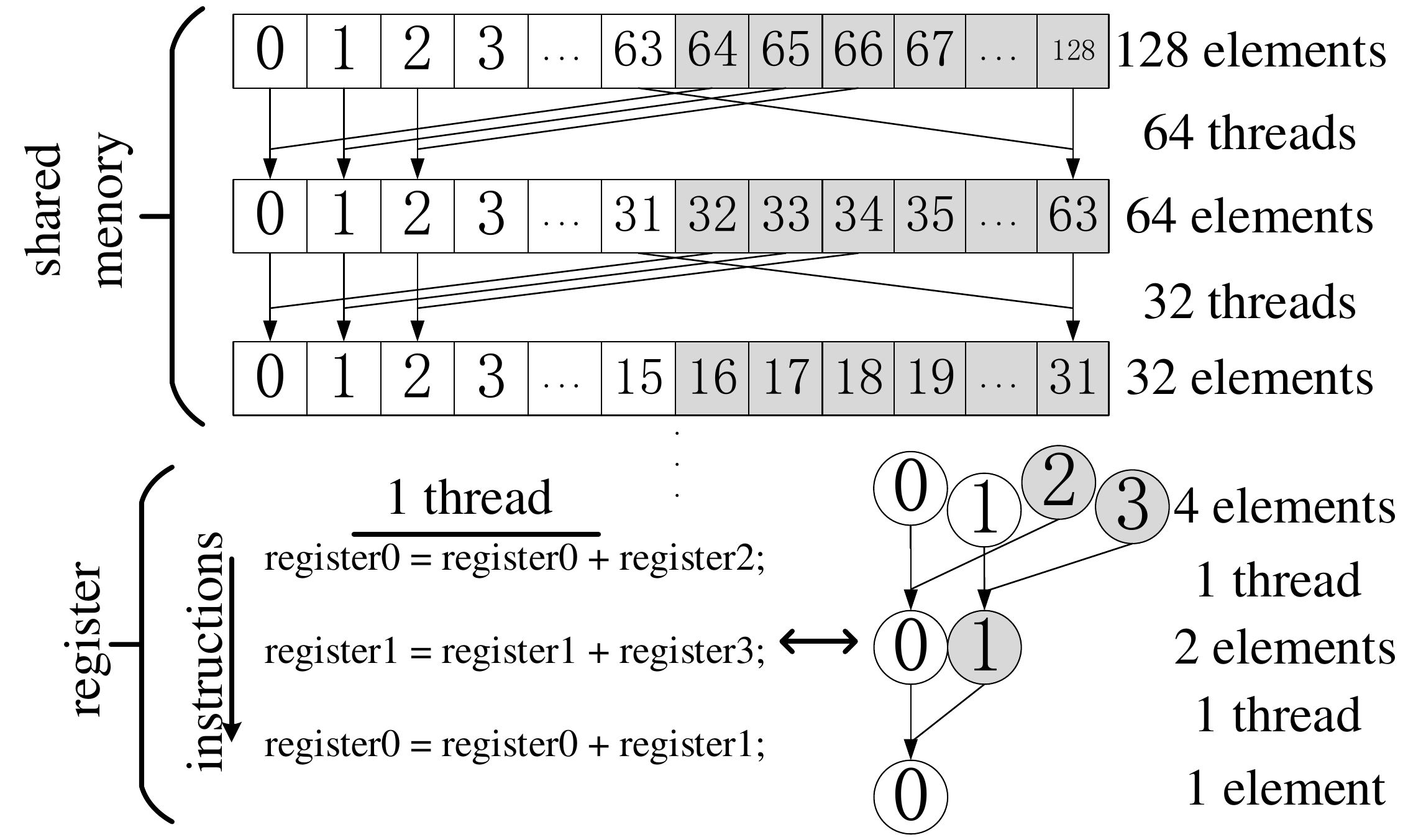}
	\caption{The diagram of improved parallel reduction on GPU}
	\label{fig:improve}
\end{figure}

The proposed method is shown in Fig.\ref{fig:improve}, the registers are used instead of the shared memory in the last $N_{r}$ rounds of the reduction. Since the number of threads is small in the last several rounds, using registers will not waste too many threads but save much shared memory access time. In Fig.\ref{fig:improve}, in the last two rounds, the classic method of parallel reduction requires to access shared memory 2 times, need to do addition 2 times. So it needs about 22 cycles to complete the operation of the last two calculations. And after the use of registers, the proposed method need to access registers 3 times, and need to do addition 3 times. So it needs about 6 cycles to complete the operation of the last two calculations. Thus the proposed method greatly improves the speed of operation. In section 4, we show the relationship between the $N_{r}$ rounds registers involved in and the time of image matching.
\subsection{Improved Parallel Euclidean Distance Calculation}
In step 4 of CasHash, calculating the Euclidean distances still spends a lot of time in CPU, and the Euclidean distance formula is $\sum _{i=0}^{d-1}\left( x_{i}^{1}-x_{i}^{2}\right) ^{2}=\Delta x\cdot \Delta x$. Therefore, the Euclidean distance is essentially a vector inner product, which is the same as that in the section 3.2. After Euclidean distance calculation and Lowe's ratio test \cite{Lowe2004}, the output (SIFT matching pairs) can be copied into hard disk. Since CPU and GPU are different devices, the operations of CPU and GPU are asynchronous \cite{Micikevicius2009}. When doing image matching on GPU, CPU can copy the output from GPU memory to host memory and then to hard disk. This method hides the output time, and improves the overall operation efficiency.
\section{Improved Parallel Hashing Ranking}
\begin{figure}
	\centering
	\includegraphics[width=70mm]{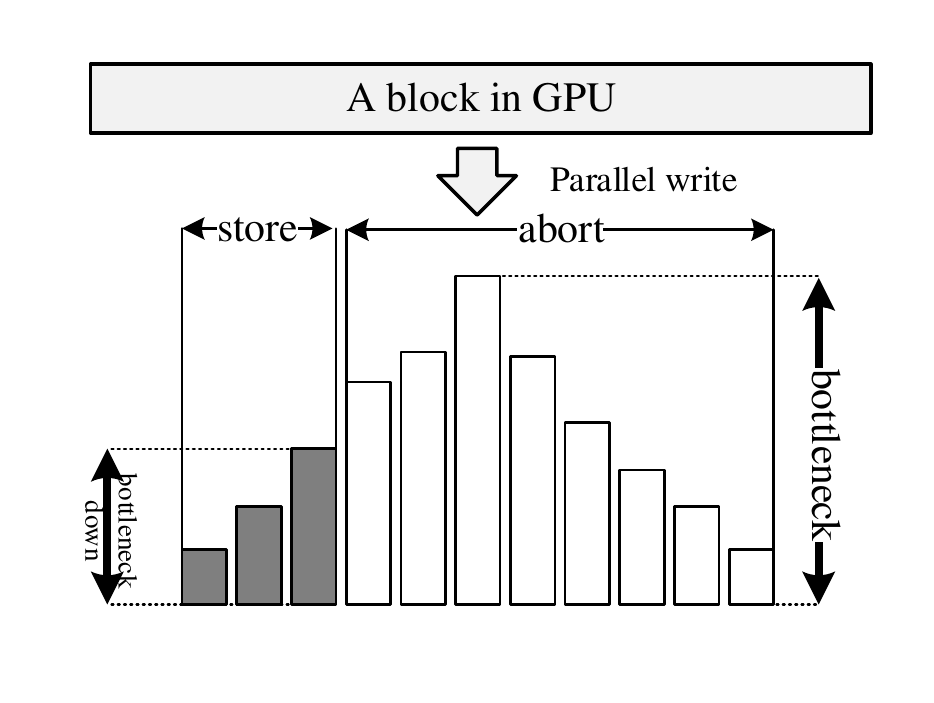}
	\caption{The diagram of the bottleneck of parallel hashing ranking. The bucket number of the grey buckets is smaller than the threshold, and the bucket number of the white buckets is bigger than the threshold. }
	\label{fig:rank1}
\end{figure}

In step 3 of CasHash, we propose an improved parallel hashing ranking method. Each GPU block ranks the Hamming distance between a query point and its candidate points in an image, and the process of constructing a index table in a GPU block is shown in Fig.\ref{fig:rank1}. As the histogram showed, the horizontal axis is the Hamming distance and the vertical axis is the number of points that fall in the bucket. On the one hand, since the candidate points have passed through rough filtering, few outliers are included. So few Hamming distances are large. On the other hand, there are $2^{128}$ kinds of different hashing codes, There are few Hamming distances near 0. Therefore in experiments, the candidate points are concentrated in the middle of the hashing bucket. If two candidate points have the same Hamming distance, there will be conflict when storing them into the same hashing bucket. In order to avoid the error caused by the conflict, it is necessary to use the atomic operation of CUDA. The atomic operation can make the conflict become serial task \cite{Sanders2010}, which will reduce the efficiency of operation. And the time of conflict in the middle of the hashing bucket is most. The efficiency of parallel computation follows the bucket principle, so the computation time depends on the points number in the middle of the hashing bucket.

However, we only need the first $k$ nearest neighbors. $k$ is less than the number of hashing buckets, so most of the $k$ nearest neighbors is in the first few buckets. In Fig.\ref{fig:rank1}, we set a threshold $\tau$ in parallel hashing ranking. The candidate points whose Hamming distances to the query are greater than the threshold $\tau$ are discarded and not stored in the bucket and the candidate points whose Hamming distances to the query are smaller than the threshold $\tau$ are stored in the bucket. Therefore the threshold $\tau$ will reduce the computational bottleneck and improve efficiency. But if the threshold $\tau$ is too small, we may filter the real matching point which will reduce matching accuracy. In this paper, the influence of the threshold on the matching accuracy and matching time is illustrated by experiments in section 4.
\section{Image Matching with Geometry-aware Cascade Hashing on GPU}
In order to further improve the matching speed, we introduce the epipolar constraints to the above mentioned CasHashGPU framework. The new method performs in two steps. In the first step, only few top scale features are matched and a GPU accelerated fundamental matrix estimation method is implemented to compute the epipolar geometry between a pair of images. In the second step, the epipolar geometry is used to guide the matching for the remaining features. The improved geometry-aware two-step feature matching method is called Ga-CasHashGPU.
\subsection{First Stage of Image Matching}
Inspired by \cite{Shah2014,Wu2013}, we use the top $20\%$ scale SIFT features to do exhaustive image matching by CasHashGPU. Since only a small portion of feature points are involved, this stage is very fast. If at least 16 matches are found between images, then we also use GPU to estimate the fundamental matrix F between the images. And if at least $2/3$ of the initial matches are inliers, the image pair will be processed in the guided matching.
\subsection{Second Stage of Guided Matching with Geometry-aware CasHashGPU }
The epipolar geometry estimated before is further used to guide the remaining matching procedure, which is called Geometry Guided CasHashGPU. It consists of the following steps. step 1. hashing lookup with multiple tables on GPU; step 2. hashing remapping on GPU; step 3. epipolar constraints to reduce the candidate points; step 4. improved parallel hashing ranking; step 5. improved parallel Euclidean distance calculation.

\begin{figure}
	\centering
	\includegraphics[width=70mm]{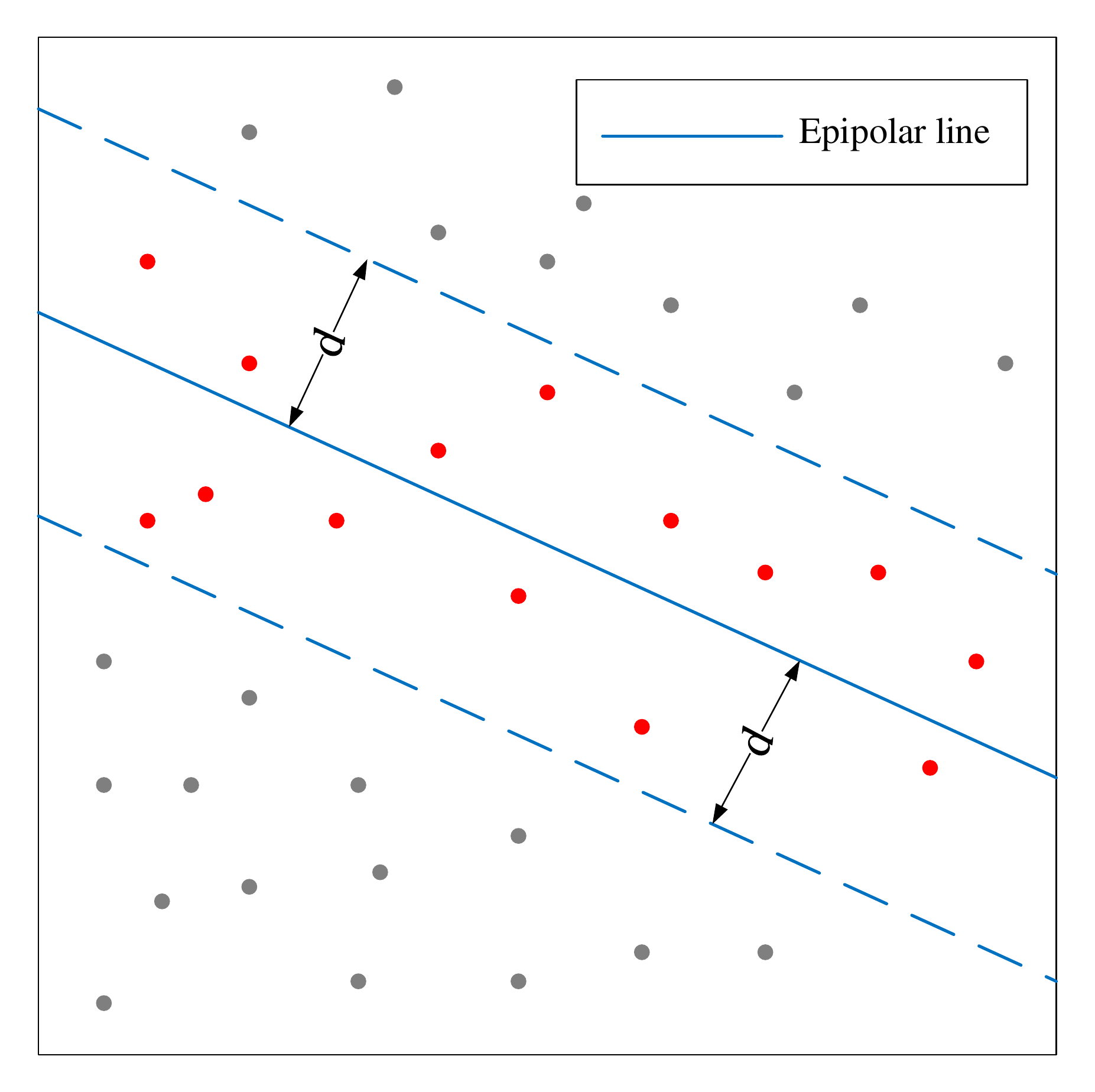}
	\caption{The diagram of the epipolar constraints. The red and gray points is the candidate points screened by hashing lookup. The red points are the points pass the epipolar constraints. }
	\label{fig:epoi}
\end{figure}

In geometry-aware feature matching \cite{Shah2015}, epipolar constraints are defined as following. $p_{q}=(x_{q}\ y_{q}\ 1)$ is a query point in image $I$, and $p'_{q}=(x'_{q}\ y'_{q}\ 1)$ is the corresponding matching point in image $J$ of $p_{q}$. $l_{q}=(a_{q},b_{q},c_{q})$ is the corresponding epipolar line in image $J$, thus the candidate matching feature set of $p_q$ $\mathbf{C}$ is,
\begin{equation}\label{candiC1}
\mathbf{C}=\{p'|dist(p',l_{q})\leq d\}
\end{equation}
\begin{equation}\label{candiC2}
dist(p',l_{q})=\dfrac {a_{q}x'+b_{q}y'+c_{q}} {\sqrt {a^{2}_{q}+b^{2}_{q}}}
\end{equation}
where $d$ is the distance of the candidate point $p'$ to the epipolar line $l_{q}$, and $d$ is the threshold of the epipolar constraints. Thus the epipolar constraints are showed in Fig.\ref{fig:epoi}. The epipolar constraints reduce the candidate points, so the proposed method reduce the points which will be sorted by improved parallel hashing ranking. Thus this way improves the overall method efficiency.

\section{Experiments and Results}
In this paper, we extracted SIFT features from the testing image data sets and perform image matching. In the experiment, the two proposed methods(CasHashGPU and Ga-CasHashGPU) are compared with Kd-Tree \cite{Muja2009}, SiftGPU \cite{SiftGPU} and the original CasHash \cite{Cheng2014}. The experiments of Kd-Tree and the original CasHash are tested on E5-2630 v3@2.4GHz CPU without parallel computation. SiftGPU and the proposed method are tested on one GTX Titan X GPU. All the experiments are running in Ubuntu 14.04 operating system. In addition, in order to test the acceleration of the proposed method on multi-GPU, we test the CasHashGPU on a PC with 8 GTX 1080Ti GPUs.
\begin{table}[]
	\centering
		\caption{Results on Matching Experiments. Kd-Tree, CasHash, Sift-GPU and CasHashGPU performs exhaustive matching without any geometric priors. Ga-CasHashGPU performs two-stage guided exhaustive matching. }
	\begin{center}
		\subtable[Data-pozzoveggiani (54 images) 8306 mean points; exhaustive matching 1431 pairs, guided matching 34 pairs]{
			\begin{tabular*}{105mm}{p{30mm}<{\centering}p{20mm}<{\centering}p{20mm}<{\centering}p{20mm}<{\centering}}
				\hline
				Method                  & time(s)   &  speed(pairs/s)& speedup\\
				\hline
				Kd-Tree                 &698.921&2.05   & 1.00$\times$  \\
				CasHash                 &183.218&7.81   & 3.81$\times$  \\
				SiftGPU                 &30.103 &47.54  & 23.22$\times$ \\
				CasHashGPU              &2.496  &573.32 & 280.02$\times$\\
				Ga-CasHashGPU           &0.736  &1990.49& 949.63$\times$\\
				\hline
			\end{tabular*}
		}
		\subtable[Data-erpbero (259 images) 8641 mean points; exhaustive matching 33411 pairs, guided matching 780 pairs]{
			\begin{tabular*}{105mm}{p{30mm}<{\centering}p{20mm}<{\centering}p{20mm}<{\centering}p{20mm}<{\centering}}
				\hline
				Method                  & time(s)   &  speed(pairs/s)    & speedup       \\
				\hline
				Kd-Tree                 &1.479$\times10^4$ &2.26   & 1.00$\times$   \\
				CasHash                 &1461.525&22.86  & 10.12$\times$  \\
				SiftGPU                 &752.164 &44.42  & 19.66$\times$  \\
				CasHashGPU              &34.394  &971.42 & 429.91$\times$ \\
				Ga-CasHashGPU           &15.701  &2177.63& 941.77$\times$ \\
				\hline
			\end{tabular*}
		}
		
		\subtable[Data-Aos\_Hus (811 images) 7768 mean points; exhaustive matching 328455 pairs, guided matching 7337 pairs]{
			\begin{tabular*}{105mm}{p{30mm}<{\centering}p{20mm}<{\centering}p{20mm}<{\centering}p{20mm}<{\centering}}
				\hline
				Method                  & time(s)   &  speed(pairs/s)      & speedup       \\
				\hline
				Kd-Tree         &1.456$\times10^5$ &2.26    & 1.00$\times$  \\
				CasHash         &2.800$\times10^4$ &11.73   & 5.20$\times$  \\
				SiftGPU         &6971.441&47.11    & 20.89$\times$ \\
				CasHashGPU      &292.541 &1122.77  & 497.81$\times$\\
				Ga-CasHashGPU   &90.189  &3723.20   & 1614.652$\times$\\
				\hline
			\end{tabular*}
		}
	\end{center}

	\label{table1}
\end{table}
\begin{table}[]
	\centering
	\caption{Results on Matching of Large Data Sets. Because Data-Dubrovnik6K and Data-Rome16K are large data sets, Kd-Tree, CasHash and SiftGPU will spend a lot of time, we only compare and our two method.}
	\begin{center}
		\subtable[Data-Dubrovnik6K \cite{Cornell} (6044 images) 7438 mean points; exhaustive matching 1.826$\times10^7$ pairs, guided matching 58611 pairs]{
			\begin{tabular*}{105mm}{p{40mm}<{\centering}p{30mm}<{\centering}p{30mm}<{\centering}}
				\hline
				Method                & time(s)  &  speedup  \\
				\hline
				CasHashGPU             &1.054$\times10^4$ & 1.00$\times$    \\
				Ga-CasHashGPU          &1548.89 & 6.80$\times$    \\
				\hline
			\end{tabular*}
		}
		\subtable[Data-Rome16K \cite{Cornell} (15178 images) 7891 mean points; exhaustive matching 1.152$\times10^8$ pairs, guided matching 145101 pairs]{
			\begin{tabular*}{105mm}{p{40mm}<{\centering}p{30mm}<{\centering}p{30mm}<{\centering}}
				\hline
				Method                & time(s)  &  speedup  \\
				\hline
				CasHashGPU             &1.565$\times10^5$ & 1.00$\times$  \\
				Ga-CasHashGPU          &20863.68 & 7.50$\times$  \\
				\hline
			\end{tabular*}
		}
	\label{table2}
\end{center}
\end{table}

Table \ref{table1} and \ref{table2} show the comparison of four methods above. The Ga-CasHashGPU performs epipolar-geometry guided exhaustive matching and other method perform exhaustive exhaustive matching without priors. CasHashGPU is about 20 times faster than SiftGPU on the same graphics card, nearly 100 times faster than the CPU CasHash method and hundreds of times faster than the CPU Kd-Tree based matching method. Ga-CasHashGPU calculates much fewer matching pairs than CasHashGPU. So Ga-CasHashGPU is about 3 to 7 times faster than CasHashGPU on the same graphics card. The experiments show that the speedup of our method is more obvious when the number of images is larger, such as the ''pozzoveggiani'' data set. This is because when the number of images increases, the scale of the problem increases quadratically.

\begin{figure}
	\centering
	\includegraphics[width=70mm]{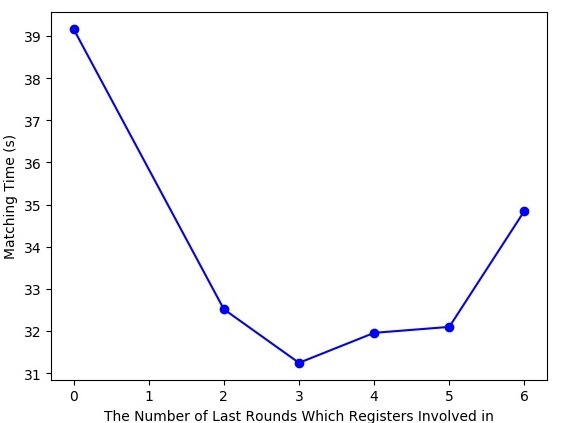}
	\caption{Results on the Improved Parallel Reduction. The horizontal axis is the last $N_{r}$ rounds registers involved in.}
	\label{fig:exp42}
\end{figure}

Fig.\ref{fig:exp42} shows the matching time on the ''erpbero'' dataset when using registers in different last $N_{r}$ rounds. When $N_{r}$ increases, the matching time will first decrease and then rise after $N_{r}$ is larger than a certain value. This is because using registers in the last few rounds will reduce memory access time, but if registers are used too early in the parallel reduction a lot of threads will be idle. We empirically set $N_{r}=3$ to achieve the best performance in our experiments.

\begin{figure}
	\centering
	\subfigure[]
	{
		\label{fig:ea}
		\includegraphics[width=58mm]{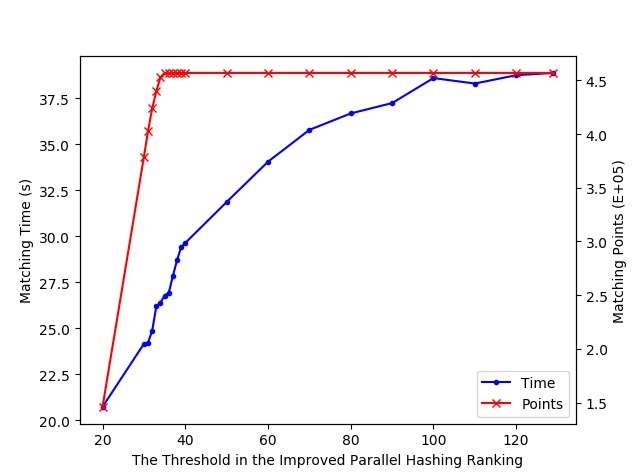}
	}
	\subfigure[]
	{
		\label{fig:eb}
		\includegraphics[width=58mm]{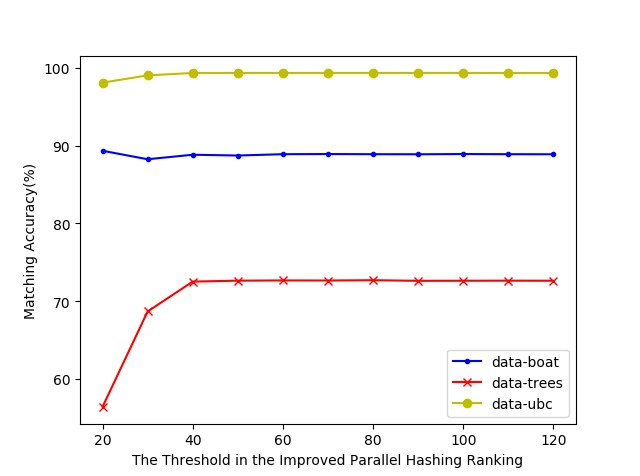}
	}
	\caption{The influence of the threshold $\tau$. The influence of the threshold $\tau$ on the matching points and matching time is showed in (a). The influence of the threshold $\tau$ on the matching accuracy is showed in (b).}
	\label{fig:exp4344}
\end{figure}

\begin{figure}
	\centering
	\subfigure[Data-Dubrovnik6K]
	{
		\label{fig:compa}
		\includegraphics[width=58mm]{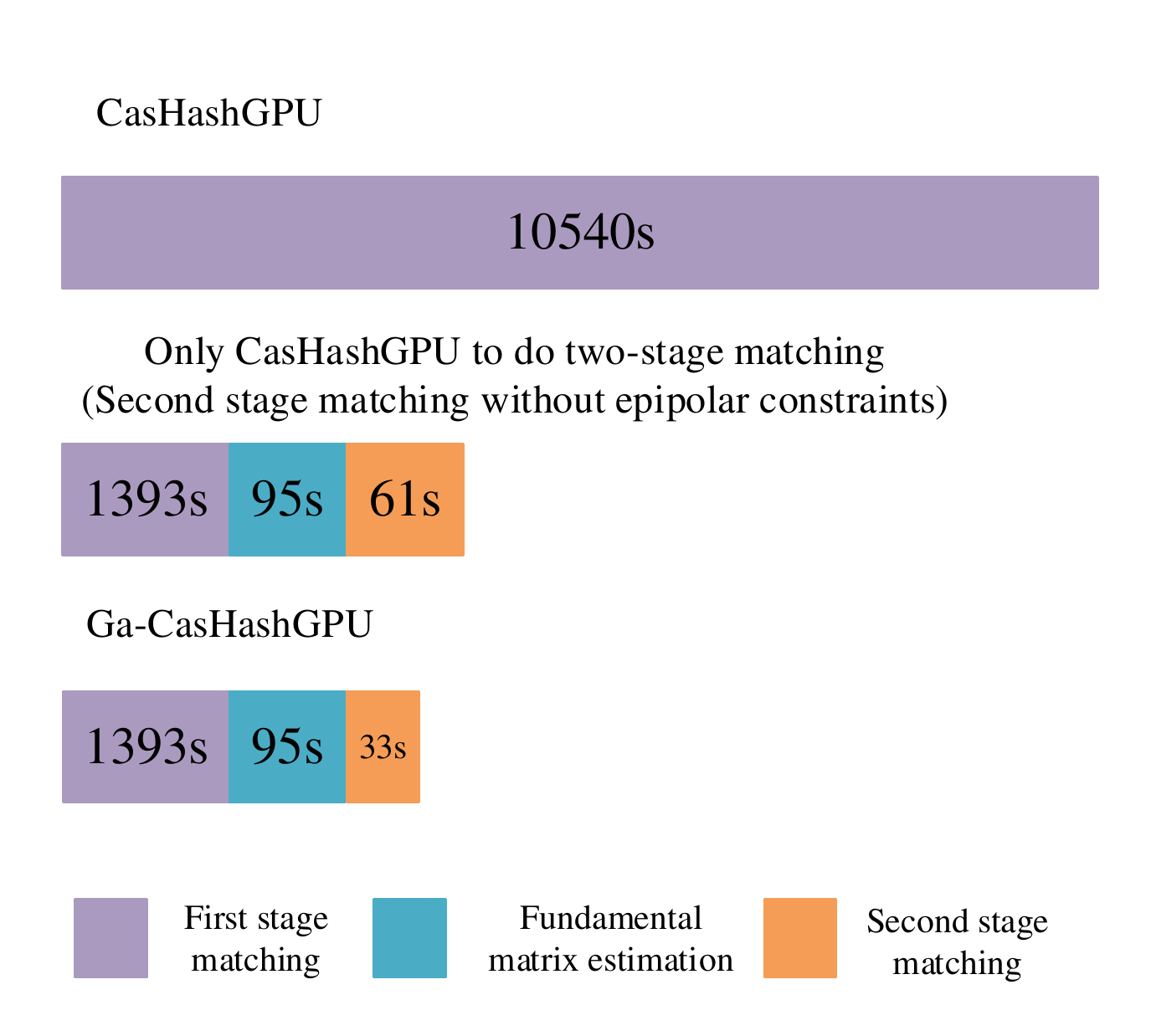}
	}
	\subfigure[Data-Aos\_Hus]
	{
		\label{fig:compb}
		\includegraphics[width=58mm]{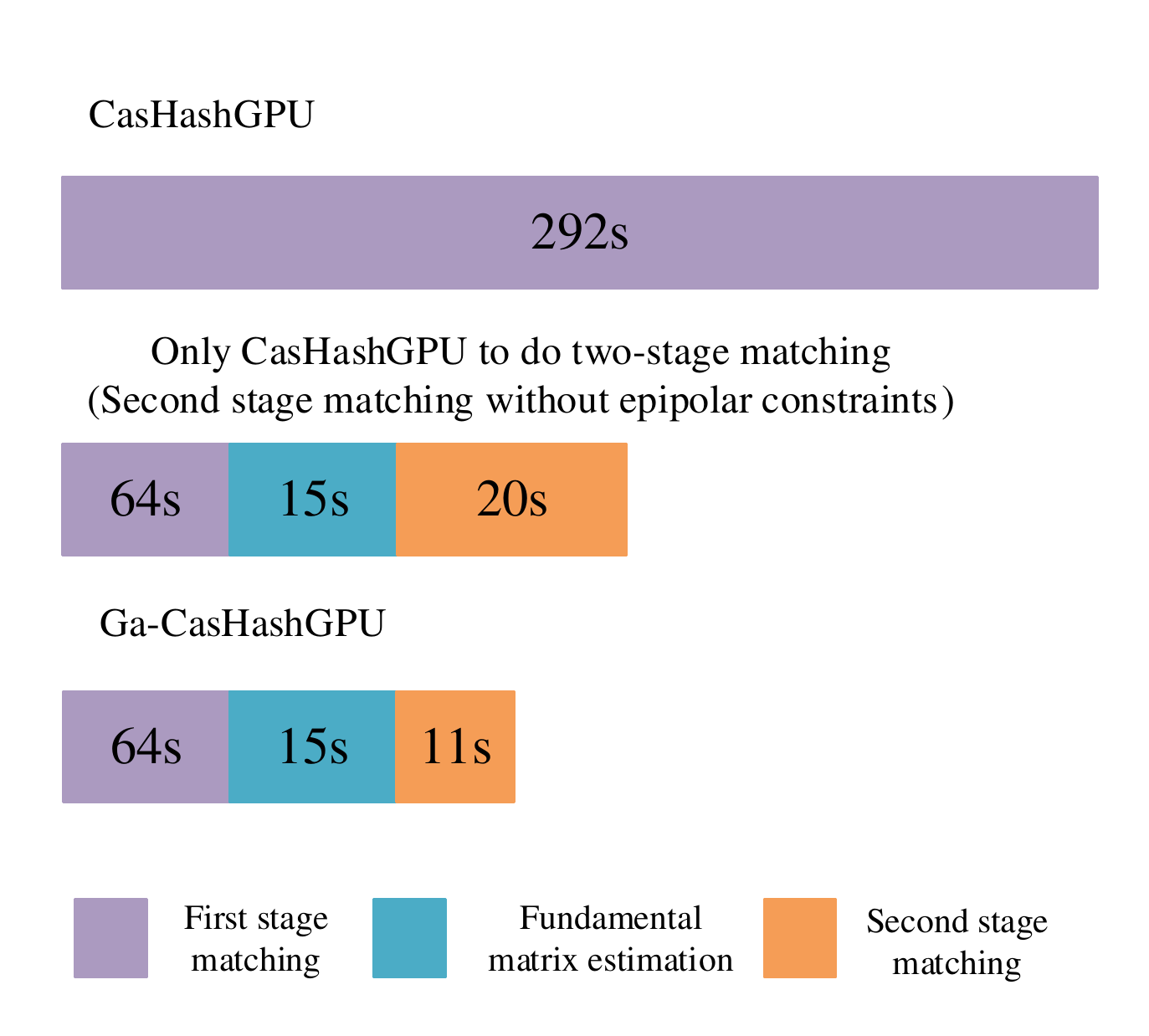}
	}
	\caption{The diagram of the analysis and comparison of time cost on ''Aos\_Hus'' and ''Data-Dubrovnik6K'' data set in each matching stage.}
	\label{fig:comp}
\end{figure}

Fig.\ref{fig:ea} shows the number of matching points and matching time on the ''erpbero'' dataset for different $\tau$. Fig.\ref{fig:eb} gives the mean matching accuracy for different $\tau$ on ''data-boat'', ''data-trees'' and ''data-ubc'' (each group use 5 image pairs) of public Oxford dataset \cite{Mikolajczyk2005}, which offers the Homography transformation as the ground truth. If the match pairs satisfy $x_{1}-Hx_{2}<\epsilon$, where $H$ is the given Homography matrix between image pairs and we set $\epsilon=6$, we assume that the two keypoints compose accurate matching pairs. In Fig.\ref{fig:exp4344}, when the threshold $\tau$ is about 40, the matching points don't decrease and we can guarantee high matching accuracy and less matching time.
Fig.\ref{fig:comp} shows the analysis and comparison of time cost on ''Aos\_Hus'' and ''Data-Dubrovnik6K'' data set in each matching stage. The experiments show that the GPU accelerated fundamental matrix estimation method is about 160 times faster than the classical RANSAC algorithm on CPU. And the epipolar constraints can make CasHashGPU faster.

\begin{figure}
	\centering
	\subfigure[]
	{
		\label{fig:ea2}
		\includegraphics[width=58mm]{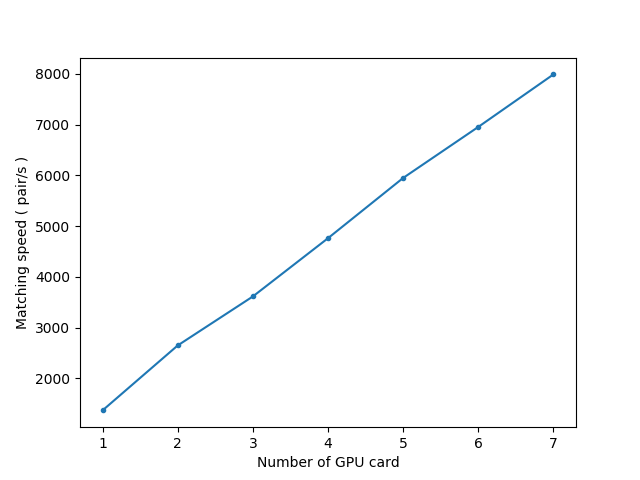}
	}
	\subfigure[]
	{
		\label{fig:eb2}
		\includegraphics[width=58mm]{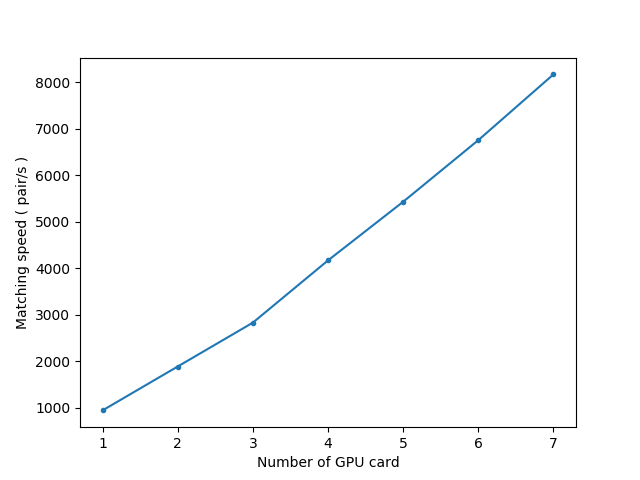}
	}
	\caption{The relationship between the number of GPU card and matching speed. The experiment on Data-Dubrovnik6K \cite{Cornell} time is showed in (a). The experiment on Data-Rome16K \cite{Cornell} time is showed in (a).}
	\label{fig:expgpu}
\end{figure}

Fig.\ref{fig:expgpu} shows the relationship between the number of GPU card and matching speed. In the experiment, the number of GPU varies from 1 to 7. In Fig.\ref{fig:expgpu}, the matching speed is linear with the number of GPUs, so CasHashGPU can linearly accelerate image matching on multi-GPU devices.

\section{Conclusions}
In this paper, we proposed a GPU accelerated image matching algorithm with CasHash for fast feature matching of massive images. We proposed an improved parallel reduction and an improved parallel hashing ranking to optimize and parallelize the Cascade Hashing. For massive images, we proposed a disk-memory-GPU data exchange strategy and we optimize the load order of image data.
\section*{References}

\bibliography{GPU_Accelerated_Cascade_Hashing_Image_Matching_for_Large_Scale_3D_reconstruction}

\end{document}